\pdfoutput=1
\documentclass[10pt,twocolumn,letterpaper]{article}

\usepackage{iccv}
\usepackage{times}
\usepackage{epsfig}
\usepackage{graphicx}
\usepackage{amsmath}
\usepackage{amssymb}

\usepackage{algorithm}
\usepackage{algorithmic}

\graphicspath{{figs/}}

\usepackage[breaklinks=true,bookmarks=false]{hyperref}

\iccvfinalcopy 

\def\iccvPaperID{****} 

\begin{document}

\title{An Improved Self-supervised GAN via Adversarial Training}

\author{Ngoc-Trung Tran, Viet-Hung Tran, Ngoc-Bao Nguyen, Ngai-Man Cheung\\
ST Electronics - SUTD Cyber Security Laboratory\\
Singapore University of Technology and Design\\
}

\maketitle

\begin{abstract}

We propose to improve unconditional Generative Adversarial Networks (GAN) by training the self-supervised learning with the adversarial process. In particular, we apply self-supervised learning via the geometric transformation on input images and assign the pseudo-labels to these transformed images. (i) In addition to the GAN task, which distinguishes data (real) versus generated (fake) samples, we train the discriminator to predict the correct pseudo-labels of real transformed samples (classification task). Importantly, we find out that simultaneously training the discriminator to classify the fake class from the pseudo-classes of real samples for the classification task will improve the discriminator and subsequently lead better guides to train generator. (ii) The generator is trained by attempting to confuse the discriminator for not only the GAN task but also the classification task. For the classification task, the generator tries to confuse the discriminator recognizing the transformation of its output as one of the real transformed classes. Especially, we exploit that when the generator creates samples that result in a similar loss (via cross-entropy) as that of the real ones, the training is more stable and the generator distribution tends to match better the data distribution. When integrating our techniques into a state-of-the-art Auto-Encoder (AE) based-GAN model, they help to significantly boost the model's performance and also establish new state-of-the-art Fr\'echet Inception Distance (FID) scores in the literature of unconditional GAN for CIFAR-10 and STL-10 datasets.

\end{abstract}

\section{Introduction}

Generative Adversarial Networks (GAN) \cite{goodfellow-nisp-2014} have become the most popular approach to train the generative model. It gets much attention from the community because of its ability to generate visual appealing samples, but not require the explicit analytic form of objective functions. The idea behind GAN is to use a binary classifier, so-called the discriminator. Discriminator learns to distinguish the data (real) versus generated (fake) samples, and as a result, it represents this manifold via its scalar scores in the form of likelihood. Training generator of GAN is to maximize discriminator's likelihood scores computed over fake samples. In other words, it confuses the discriminator to accept its outputs as the real ones. Training GAN is an the adversarial process, in which the discriminator and generator compete with each other to improve themselves. Although GAN is an attractive approach, using the real/fake label to train GAN is challenging because this supervisory signal is a weak constraint. Hence, the generator can easily cheat the discriminator by, eg., always creating the identical samples but recognized with high likelihood by the discriminator. It explains why GAN has many serious issues, such as the gradient vanishing and mode collapse \cite{goodfellow-nips-2016,arjovsky-arxiv-2017a}, which prevent the model to possibly cover all modes of the data distribution. Many variants of GAN have proposed new constraints to overcome this ill-pose problem.

In the literature, many constraints have been proposed for the discriminator. These constraints force discriminator's gradients not to be vanishing so that the generator can use them to learn and improve itself. Intuitively, these constraints smoothen out the decision boundary of the discriminator between real and fake samples in order to avoid the sharp gradients along this region and enable distant samples to contribute more to generator training. One of the most noticeable regularization techniques are towards enforcing Lipschitz conditions \cite{Arjovsky2017,gulrajani-arxiv-2017,roth-nips-2017,kodali-arxiv-2017,petzka-arxiv-2017,liu-arxiv-2018,miyato-iclr-2018}. However, these techniques have their own disadvantages, for example, the divergence issue \cite{zhang-arxiv-2018} as the regularization becomes over-strength at the end. Overcoming this requires careful designs of training procedure \cite{yazici-arxiv-2018,heusel-arxiv-2017}.

The alternative constraints, which are also commonly-used, are via auto-encoder. It reconstructs the real samples, hence guides the generator to produce samples resembling the real modes. It increases the chance to occur the competition between discriminator and generator on many modes of data distribution. Therefore, it potentially encourages the discriminator can create better gradients that lead to a better generator. However, the downside of auto-encoder is the blurry issue. Although some recent works \cite{larsen-arxiv-2015,tran-eccv-2018} overcame this problem by using the high-level features of discriminators, the texture and shape of objects in generated images does not look realistic.

A recent GAN \cite{chen-arxiv-2018} proposed new constraints via self-supervised learning strategy \cite{gidaris-iclr-2018}. The authors argument new samples via image rotation and assign them with pseudo-labels. In addition to training discriminator to distinguish real and fake samples, they train discriminator to predict the correct labels of rotated images. They train the generator to minimize the classification loss as the discriminator recognizes the transformation of generator's outputs. In other words, they train the generator to create images whose correct pseudo-labels of their transformed samples are easily recognized by the discriminator. Although its results are encouraging, the discriminator does not take into account the generated samples for classification task and it's not precise how self-supervised tasks helped to improve GAN in this work. In fact, the proposed generator objective \cite{chen-arxiv-2018} minimizes the cross-entropy loss, which does not necessarily help to create samples resembling real samples. For example, like original GAN, the generator may create collapsed samples, but recognized as real by the discriminator with high probability and its rotated samples are still classified correctly according to their pseudo-label ground-truth.

In this work, we propose an improved self-supervised GAN, which introduces the adversarial way of using self-supervised learning. In particular, we first propose to train discriminator to classify correct pseudo-labels of real transformed samples (obtained from data samples via geometric transformation) as the classification task. This classification task improves GAN model when being combined with the original GAN task \cite{goodfellow-nisp-2014} that learns to distinguish data (real) versus generated (fake) samples. Then, we propose two further improvements: (i) We propose to train the discriminator to simultaneously classify the class of generated samples from pseudo-classes of real samples. We consider it as the adversarial training for the discriminator. This adversarial training significantly improves the discriminator, hence improves the generator and the model performance. (ii) In addition to confusing the GAN task, we propose a new generator objective to fool the classification task of the discriminator by creating samples that the discriminator recognizes their transformed ones as real pseudo-classes. Importantly, instead of minimizing the cross-entropy of transformed fake samples like the previous work, we do match the cross-entropy loss computed over fake transformed samples to that of the real transformed ones. We exploit that it stabilizes the training, boosts the significantly performance as being combined with adversarial training of discriminator. We investigate our proposed techniques with the state-of-the-art AE-based GAN model \cite{tran-eccv-2018}. Although \cite{tran-eccv-2018} demonstrated that the combination of auto-encoder and gradient penalty constraints combined together improve the training of GAN and achieve state-of-the-art performance, integrating our techniques can further boost the performance of this baseline model. We see that benefiting all kind of constraints in a good way will stabilize GAN, and establish a new state-of-the-art performance on CIFAR-10 and STL-10 datasets.
\section{Related Work}

While training GAN with conditional signals (e.g., class labels) \cite{odena-icml-2017,zhang-arxiv-2018,brock-iclr-2018} are attaining promising results, training GAN in the unconditional setting is still challenging. In the original GAN \cite{goodfellow-nisp-2014}, the single signal (real or fake) of samples are provided to train discriminator and use the discriminator to guide the generator. With these signals, the generator or discriminator may fall into ill-pose settings, where easily being stuck at bad local minimums though still satisfying the signal constraints. Therefore, many regularizations have been proposed to reduce this problem, and the most popular technique is to enforce (or towards) Lipschitz condition of the discriminator by weight-clipping \cite{arjovsky-arxiv-2017a}, gradient penalty constraints \cite{gulrajani-arxiv-2017,roth-nips-2017,kodali-arxiv-2017,petzka-arxiv-2017,liu-arxiv-2018}, consensus constraint, \cite{mescheder-nips-2017,mescheder-icml-2018}, or spectral norm \cite{miyato-iclr-2018}. Constraining the discriminator in such ways to prevent its gradients vanishing, and avoid the sharp boundary decision between real and fake classes. Otherwise, because the data points are very sparse in a high-dimensional manifold, without strong constraints, the discriminator is able to always find the perfect decision boundary between real and generated data points as it is powerful enough. It is likely the main reason causing the gradient vanishing issues of GAN.

Although regularizations improve the stability of GAN, using a single supervisory signal like original GAN \cite{goodfellow-nisp-2014} still leads to challenging optimization problems. It is because that discriminator scores are highly dependent on generated samples. Therefore, if the generator is collapsed to some particular modes of data distribution, it is only able to create samples around these modes. Subsequently, there is no competition to train the discriminator around other modes. As a result, the gradients of these modes may be vanishing, and it is impossible to guide the generator to model the entire data distribution. Using more supervisory signals simplifies the optimization process. For example, using self-supervised learning in the form of auto-encoder. AAE \cite{makhzani-arxiv-2015} guides the generator towards creating more realistic samples. It is a potential solution to partly prevents the generator from generating identical samples. It steers the generated samples towards real samples to reduce the disjoint issue between two distributions, therefore, less be over-fitting and gradient vanishing. However, the problem of using auto-encoder is that pixel-wise reconstruction with $\ell_2$-norm would cause the blurry issue. VAE/GAN \cite{larsen-arxiv-2015}, which combined  VAE \cite{kingma-arxiv-2013} and GAN, suggest a better solution: while the discriminator of GAN enables the usage of feature-wise reconstruction to overcome the blur, the VAE constrains the generator better to reduce the mode collapse. ALI \cite{dumoulin-arxiv-2016} and BiGAN \cite{donahue-arxiv-2016} jointly train the data/latent samples in GAN framework like to put more constraints on the discriminator and the generator. InfoGAN \cite{chen-arxiv-2016} infer the disentangled representation of latent code by maximizing the mutual information. In addition to using feature-wise, \cite{tran-eccv-2018,tran-aaai-2018} combine the two different types of supervisory signals: real/fake signals and self-supervised signal in the form of auto-encoder, which lead to stable convergence and better-generated images and prevent the model from the mode collapse. Although feature-wise distance for auto-encoder is often good to reconstruct the sharper images, its reconstructed images still cannot produce realistic detail of textures or shapes.

Recently, self-supervised learning is getting much attention from the community as it helps to close the gap between supervised and unsupervised models in classification tasks \cite{doersch-cvpr-2015,pathak-cvpr-2016,zhang-eccv-2016,zhang-cvpr-2017,noroozi-iccv-2017,gidaris-iclr-2018}. This technique encourages the classifiers to learn better feature representation with pseudo-labels, which has been also applied for GAN \cite{chen-arxiv-2018}. However, the usage of the self-supervised task in this work is simply following the idea of \cite{gidaris-iclr-2018}. It's unclear how the classification tasks help the model. Moreover, although the usage of self-supervised learning to train discriminator is simple, making use of self-supervised learning effective for the generator is not trivial.

\section{Proposed Method}

\begin{figure*}[htb!]
\centering
\includegraphics[scale=0.44]{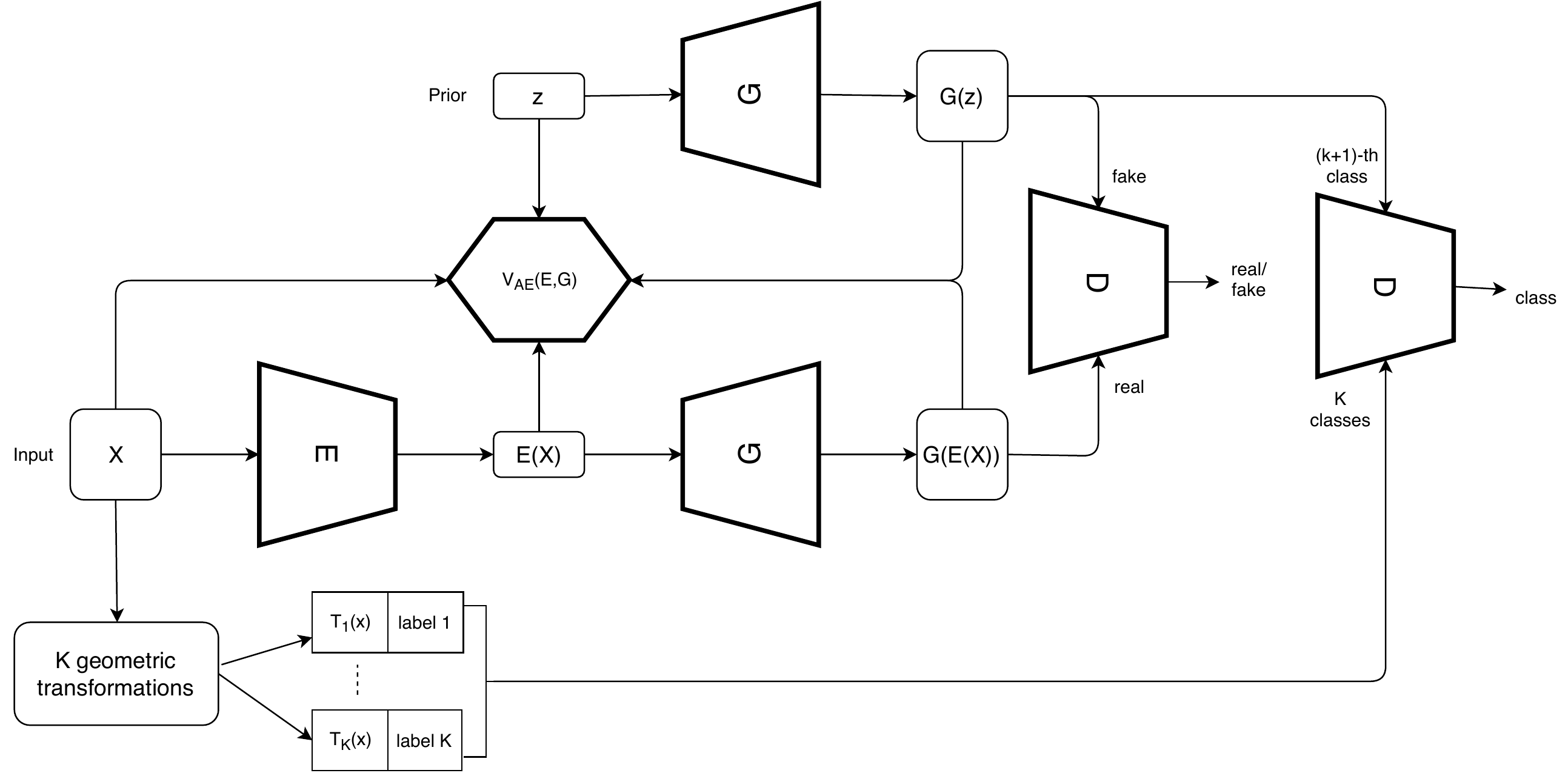}
\caption{The diagram of our model. $E$, $G$, $D$ are the encoder, the generator, and the discriminator. The parameters of $G$ are shared for the generator and decoder of auto-encoder. Two discriminators ($D$) are shared parameters excepts two different heads: one dimension of the real/fake classes and $K+1$ dimensions of pseudo-classes of geometric transformation. The real image ($X$) is encoded and decoded into the reconstruction ($G(E(X))$). Here, we show the reconstruction for clarification, in our implementation we use the features of discriminator $\Phi(X)$. The constraint $V_R(E,G)$ is to regularize the reconstruction like \cite{tran-eccv-2018}. The construction is considered as the ``real" sample when optimizing the discriminator with the objective $\mathcal{V}_{D_{gan}}$. The input $X$ is transformed into $K$ new samples with their pseudo-labels, and the discriminator $D$ is trained to recognize the correct labels, and also to classify the fake samples from the K real classes.}
\label{gan-model}
\end{figure*}

In our work, we adopt an auto-encoder based method, Dist-GAN \cite{tran-eccv-2018}, to be our baseline model because it has already demonstrated the combination of gradient penalty and auto-encoder constraints achieves the state-of-the art-results of GAN. We discuss adversarial self-supervised learning (in short of training self-supervised learning with the adversarial process) for the discriminator and the generator and how to integrate them into the baseline model. Our model consists of three main components: we use the regularized auto-encoder (consisting of the encoder (E) and decoder (G)) like \cite{tran-eccv-2018}, and we propose new objectives of the discriminator (D) and the generator (G) to improve the model. The decoder and the generator share all parameters. In our model, we first train the auto-encoder, after that we train the discriminator to distinguish real and fake samples (GAN task) and also learn to predict correct augmented labels (classification task) and finally we train the generator to match real and fake scores in combination with matching the cross-entropy losses computed over transformations of these samples. Our components and the training algorithm are represented in Fig. \ref{gan-model} and Alg. \ref{alg-01}. To highlight our main contributions, we will first discuss our proposed discriminator and generator objectives and then remind the regularized auto-encoder.

\begin{algorithm}
 \footnotesize
 \caption{Our training algorithm}
 \begin{algorithmic}[1]
 \STATE Initializing parameters of discriminator, encoder and generator $D, E, G$ respectively. $N_{iter}$ is the number of iterations.
 \REPEAT
  \STATE $\mathrm{x} \leftarrow$ Randomizing mini-batch of $N$ samples from dataset.
  \STATE $\mathrm{x}^l \leftarrow$ Argument-ing samples $\mathrm{x}$ by image transformation task $l$.
  \STATE $\mathrm{z} \leftarrow$ Randomizing $N$ samples from noise distribution $P_\mathrm{z}$
  \STATE // \textit{Training the auto-encoder using $\mathrm{x}$ and $\mathrm{z}$ according to Eqn. \ref{eq-regularized-autoencoder}}
  \STATE $E, G \leftarrow \min \mathcal{V}_{AE}(E,G)$
  \STATE // \textit{Training discriminator/classifier according to Eqn. \ref{eq_D_obj} on $\mathrm{x}, \mathrm{x}^l, \mathrm{z}$} 
  \STATE $D \leftarrow \max \mathcal{V}_D(D,G)$  
  \STATE // \textit{Training the generator on $\mathrm{x}, \mathrm{z}$  according to Eqn. \ref{eq:genobj_new}.} 
  \STATE $G \leftarrow \min \mathcal{V}_G(D,G)$  
 \UNTIL $N_{iter}$
 \RETURN $D, E, G$ 
 \end{algorithmic} 
 \label{alg-01}
\end{algorithm}

\subsection{Discriminator Objective}
\label{discriminator-objective}

Our discriminator objective (Eq. \ref{eq_D_obj}) consists of two parts: (i) The GAN objective to train discriminator to distinguish between real/fake samples $\mathcal{V}_{D_{gan}}(D,G)$. (ii) The classification objective to train the classifier to predict the correct labels of the argument-ed samples via image transformations, $\mathcal{V}_{D_{cls}}(D)$. \textit{The discriminator and classifier are the same (shared parameters), excepts two different heads: the last fully-connected layer which returns $1$ dimension (real or fake) for the discriminator and the other returns $K$ dimensions of pseudo-classes for the classifier respectively}. $\lambda_d$ is the constant selected through empirical experiments.

\begin{equation}
\mathcal{V}_D(D,G) = \mathcal{V}_{D_{gan}}(D,G) + \lambda_d \mathcal{V}_{D_{cls}}(D)
\label{eq_D_obj}
\end{equation}

\subsubsection{GAN-based Objective}

The discriminator part for GAN is written in Eq. \ref{eq_D_gan_obj}. It's different from GAN objective \cite{goodfellow-nisp-2014} that our model considers the reconstructed samples as ``real'' represented by the term $V_C = \mathbb{E}_{\mathrm{x}} \log(D(G(E(\mathrm{x})))$, so that the gradients from discriminator are not saturated too quickly. This constraint slows down the convergence of discriminator and couples the convergence between discriminator and auto-encoder. It's likely another regularization technique, which has the similar goal as \cite{arjovsky-arxiv-2017}, \cite{miyato-iclr-2018} and \cite{tran-eccv-2018,tran-aaai-2018}.
In our method, we use a small weight for $V_C$ with $\alpha = 0.05$ for the discriminator objective. We observe that $V_C$ is important at the beginning of training. However, towards the end, especially for complex image datasets, the reconstructed samples are less useful as it may result in lower quality than the real samples. We also observe that after $N_{decay} = 150K$ training iterations, most of models does not much significantly improve the quality of images when continue the training with the same $\alpha$. From this point, we start to decay the value of $\alpha$ according to the iterations $\alpha_{decay} = \alpha * \frac{n}{N_{iter} - N_{decay}}$, where $n$ start to be counted from $N_{decay}$. Here, $\mathbb{E}$ is the expectation, $\mathbb{E}_{\mathrm{x} \sim p_d}$ and $\mathbb{E}_{\mathrm{z} \sim p_z}$ may be written as $\mathbb{E}_{\mathrm{x}}$ and $\mathbb{E}_{\mathrm{z}}$ respectively for short. $p_d$ and $p_z$ are data distribution and prior noise distribution. $\lambda_{\mathrm{p}}$ is a constant, $V_P = \mathbb{E}_\mathrm{x}(||\nabla_{\hat{\mathrm{x}}} D(\hat{\mathrm{x}})|| - 1)^2$ and $\hat{\mathrm{x}} = \mu \mathrm{x} + (1 - \mu) G(\mathrm{z})$, $\mu$ is a uniform random number $\mu \in U[0,1]$. $V_P$  enforces sufficient gradients from the discriminator to train the generator. For hinge loss, replacing $\log(D(\mathrm{x}))$ by $\min(0, -1 + D(\mathrm{x}))$ in Eq. \ref{eq_D_gan_obj}.

\begin{figure*}
\begin{equation}
\mathcal{V}_{D_{gan}}(D,G) = -((1 - \alpha)\mathbb{E}_{\mathrm{x}} \log(D(\mathrm{x})) + \alpha \mathbb{E}_{\mathrm{x}} \log(D(G(E(\mathrm{x}))) - \mathbb{E}_{\mathrm{z}} \log(1 - D(G(\mathrm{z}))) - \lambda_{\mathrm{p}}\mathbb{E}_\mathrm{x}(||\nabla_{\hat{\mathrm{x}}} D(\hat{\mathrm{x}})|| - 1)^2)
\label{eq_D_gan_obj}
\end{equation}
\end{figure*}

\subsubsection{Classification-based Objective}

The second part of the discriminator objective is for the classification task. We apply the self-supervised learning techniques to argument samples with geometric transformations and train the discriminator $D$ to predict correct pseudo-labels of these samples. In particular, we apply $K$ geometric transformations $T_k \in \mathcal{T}$ on original input $\mathrm{x}$ to create new $K$ samples $T_k(\mathrm{x})$, and assign the transformed $T_k(\mathrm{x})$ with pseudo-labels $\mathrm{y} = k$. We consider these argument-ed data samples are real transformation classes (from $1$-st to $K$-th classes), and simultaneously the generated samples are the fake transformation class ($(K + 1)$-th class). In order to train D as the multi-class classifier, we add another head into $D$ in addition to the conventional real/fake output. It is a fully-connected (FC) layer with $K+1$ soft-max outputs. Therefore, the discriminator can be also called as a classifier in this case. The goal in this section is to train the classifier to predict the geometric transformation applied to the image. We train the classifier $D$ to distinguish the $K$ real classes and fake class by minimizing the objective of Eq. \ref{obj_classification}. Note that we do not rotate the generated samples when training the discriminator, because enforcing the discriminator to recognize the correct classes of transformed fake samples makes the discriminator getting worse. It's due to that the generated samples themselves can be very noisy, especially at the beginning of the training. In addition, it seems to have some overlapping between GAN task and classification task because they both learn to classify the fake samples, however, it is important to have both tasks because each task may have its responsibility. GAN task is to distinguish between real and fake samples to approximate the distribution and classification task is to learn the useful feature representation to improve the first one. Indeed, If one of them is removed, the performance gets significantly worse. It's also worth noting that \cite{chen-arxiv-2018} only proposes the first term of our objective (Eq. \ref{obj_classification}) and does not get benefits of generated samples in the training.

\begin{figure*}
\begin{equation}
\mathcal{V}_{D_{cls}}(D) = - \mathbb{E}_{\mathrm{x} \sim p_{d}, T_k \sim \mathcal{T}}\log(P_D(\mathrm{y} = k \leq K |T_k(\mathrm{x}))) - \mathbb{E}_{\mathrm{z} \sim p_{z}}\log(P_D(\mathrm{y} = K + 1|G(\mathrm{z})))
\label{obj_classification}
\end{equation}
\end{figure*}

Here, $P_D(\mathrm{y} = k | T_k(\mathrm{x}))$ is the soft-max predicted probability of $k$-th class on data sample $\mathrm{x}$ which is transformed by $T_k$. Training the classifier to predict the pseudo-labels of real transformation classes encourages the discriminator to learn the useful feature representation of images and therefore leads to a better decision as distinguishing the real and fake samples. In addition, we train the classifier simultaneously distinguish with the fake transformation class, which is a type of adversarial training like original GAN \cite{goodfellow-nisp-2014}. The classifier learns to recognize the fake samples from the pseudo-classes of real ones is probably to create better gradients to guide the generator. Here, it's an adversarial training because there is a competition between discriminator and generator for the classification task. It's an important finding of our work, which is helpful to further improve the baseline model. It's worth noting that when we discuss adversarial training in our work, we would mean for self-supervised learning (classification task). The adversarial training for GAN task is a default. In our model, the well-trained discriminator/classifier also produces good feature-wise distance for the reconstruction task (Section \ref{regularized-auto-encoder}) to train better auto-encoder for our model because we're using discriminator features to form the reconstruction objective. It was shown in previous experiments \cite{tran-eccv-2018} that on synthetic data as the reconstruction is nearly perfect, this auto-encoder based model can approximate well the data distribution. We constrain the discriminator by the reconstruction; therefore, if the higher-quality reconstruction leads to better quality and convergence of discriminator and hence generated samples are more realistic.

\subsection{Generator Objective}
\label{generator-objective}

A recent work \cite{chen-arxiv-2018} proposed a way to integrate the self-supervised technique into GAN via image rotations \cite{gidaris-iclr-2018}. However, it is unclear how much the discriminator and generator contribute to these improvements. Not mentioning that this technique is not always applicable to other GAN methods. For example, using this self-supervised technique \cite{chen-arxiv-2018} to our generator causes our model diverged and reduces the quality of generated images (Section \ref{parameters-selection}).

\begin{equation}
\mathcal{V}_G(D,G) = \mathcal{V}_{G_{gan}}(D,G) + \lambda_g \mathcal{V}_{G_{cls}}(G)
\label{eq:genobj_new}
\end{equation}

In this work, we propose a new generator objective (Eq. \ref{eq:genobj_new}) including two terms. The first term is the GAN task $\mathcal{V}_{G_{gan}}(D,G)$, which is motivated from \cite{tran-eccv-2018} as shown in (Eq. \ref{eq:genobj_gan}). The intuition of this term is that the discriminator can model the data manifold by its scalar values. To approximate the data distribution in general, we match the two manifolds together. However, it's challenging due to high-dimensions. Therefore, we indirectly align the distribution of real discriminator scores to the distributions of generated discriminator scores. 

\begin{equation}
\mathcal{V}_{G_{gan}}(D,G) = ||\mathbb{E}_{\mathrm{x}} D(\mathrm{x}) - \mathbb{E}_{\mathrm{z}} D(G(\mathrm{z}))||
\label{eq:genobj_gan}
\end{equation}

The second term is the classification task, $\mathcal{V}_{G_{cls}}(G)$. In \cite{chen-arxiv-2018}, the generator aims to create samples $G(\mathrm{z})$ that the discriminator can easily predict its pseudo-labels for the transformed sample $T_k(G(\mathrm{z}))$. In contrast, our term is to match the self-supervised tasks to train the generator. Our intuition is that if generator distribution is similar to the real distribution, the classification performance on its transformed samples should be similar to that of those transformed from real samples. In other words, if real and fake samples are from similar distributions, the same tasks applied for real and fake samples should have resulted in similar behaviors. In particular, given the cross-entropy loss computed on real samples, we train the generator to create samples that are able to match this loss. We form the cross-entropy loss of multi-class classification as shown in Eq. \ref{entropy_matching_loss}. Here, we train the generator to confuse the classifier to recognize fake transformed samples as the same performance as it recognizes transformed classes obtained from the real ones. When the classifier learns to distinguish the real versus fake transformation classes, it learns to create good gradients and the generator gets benefits of these gradients to learn and confuse the classifiers. This adversarial process is similar to original GAN \cite{goodfellow-nisp-2014}, but now applied for multi-classes. Here, $\lambda_g$ is a constant selected through empirical experiments, and we use $\ell_1$-norm for both the GAN task and the classification task. In our implementation, we randomly select a geometric transformation $T_k$ for each data sample when training the discriminator. And the same $T_k$ are applied for generated samples when matching the self-supervised tasks to train the generator.

\begin{figure*}
\begin{equation}
\mathcal{V}_{G_{cls}}(G) = ||\mathbb{E}_{\mathrm{x} \sim p_{d}, T_k \sim \mathcal{T}}\log(P_D(\mathrm{y} = k \leq K |T_k(\mathrm{x}))) - \mathbb{E}_{\mathrm{z} \sim p_{z}, T_k \sim \mathcal{T}}\log(P_D(\mathrm{y} = k \leq K|T_k(G(\mathrm{z}))))||
\label{entropy_matching_loss}
\end{equation}
\end{figure*}

\subsection{Regularized Auto-encoder}
\label{regularized-auto-encoder}

We use the regularized auto-encoder (AE) in our model to prevent the generator from being severely collapsed and guide the generator in producing samples that resemble real ones as shown in recent works \cite{tran-eccv-2018,tran-aaai-2018}. We propose to use the similar auto-encoder objective function \cite{tran-eccv-2018}:

\begin{equation}
V_{AE}(E,G) = ||\Phi(\mathrm{x}) - \Phi(G(E(\mathrm{x})))||^2 + \lambda_\mathrm{r} V_R(E,G)
\label{eq-regularized-autoencoder}
\end{equation}

Eq. \ref{eq-regularized-autoencoder} is the objective of our regularized AE. The first term is reconstruction error in conventional AE. The second term $V_R(E,G)$ is the distance constraint, similar to \cite{tran-eccv-2018}, to regularize the mapping from latent to data samples. Here, $G$ is GAN  generator (decoder in AE), $E$ is the encoder and the constant $\lambda_\mathrm{r} = 1.0$ as suggested by the original work. $\Phi(\mathrm{x})$ is the features of the sample $\mathrm{x}$ computed through the last convolution layer of the discriminator $D$, $d_\mathrm{z} = 128$ is the dimension of latent samples $\mathrm{z}$. Here, we re-use parameters of auto-encoder from the original model and focus the analysis on our main contributions as discussed in previous sections (\ref{discriminator-objective}, \ref{generator-objective}).

\section{Experimental Results}

We conduct experiments to investigate the effectiveness of our proposed adversarial self-supervised learning on CIFAR-10 and STL-10 datasets. Images of STL-10 are resized into $48 \times 48$ like \cite{miyato-iclr-2018}. We use DCGAN \cite{radford-arxiv-2015} architecture with standard ``log" loss, and SN-GAN \cite{miyato-iclr-2018} and ResNet \cite{gulrajani-arxiv-2017} architectures with ``hinge" loss. We use ``hinge" loss for SN-GAN and ResNet because it attains better performance than standard ``log" loss as shown in \cite{miyato-iclr-2018}. We remind these networks in the supplementary material. In our model, the encoder network is the mirror of the generator network. We measure the diversity and quality of generated samples via the Fr\'echet Inception Distance (FID) \cite{heusel-arxiv-2017}. FID is computed with 10K real samples and 5K generated samples like SN-GAN \cite{miyato-iclr-2018} if not precisely mentioned. FID is computed every 10K iterations in training and visualized with the smoothening windows of 5. We train our method with 300K iterations, and report the FID of the last iteration excepts the standard SN-GAN for CIFAR-10 where we report it at about 120K because continuing the training does not improve the FID. We conduct the ablation studies and fine-tuning parameters on DCGAN, SN-GAN and ResNet architectures, and will use their best settings to compare to the state-of-the-art methods. Dist-GAN \cite{tran-eccv-2018} is our main baseline in ablation studies. We train models using Adam optimizer with learning rate $\mathrm{lr} = 0.0002$, $\beta_1 = 0.5$, $\beta_2 = 0.9$ for DCGAN and SN-GAN architectures and $\beta_1 = 0.0$, $\beta_2 = 0.9$ for ResNet architecture \cite{gulrajani-arxiv-2017}. We set $\lambda_{\mathrm{p}} = 1.0$, latent dimension is $d_\mathrm{z} = 128$ and mini-batch size is 64 for our all experiments.

\subsection{Ablation Study}
\label{parameters-selection}

\begin{figure*}
\centering
\includegraphics[width=4.8cm,height=4.0cm]{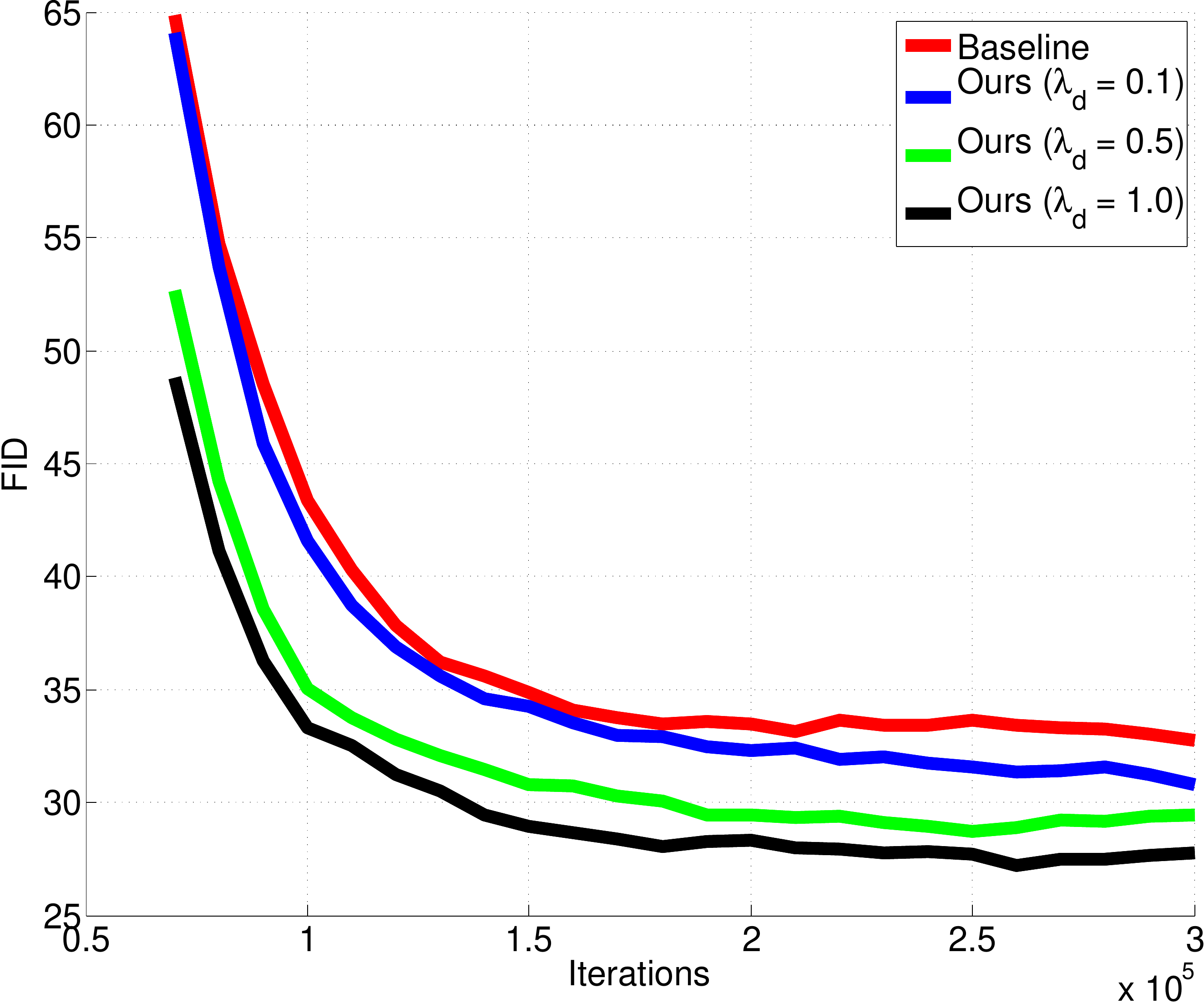}
\includegraphics[width=4.8cm,height=4.0cm]{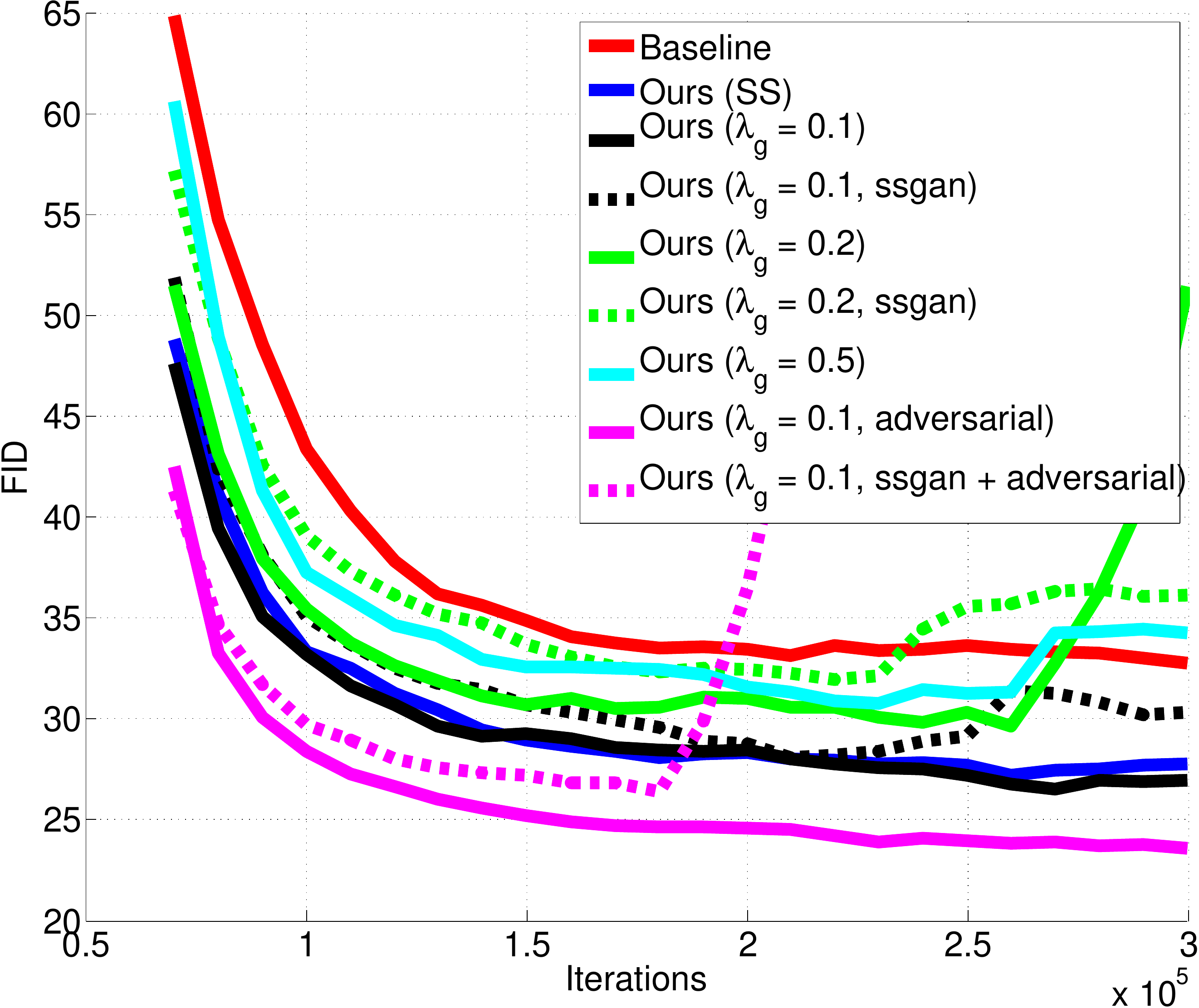}
\includegraphics[width=4.8cm,height=4.0cm]{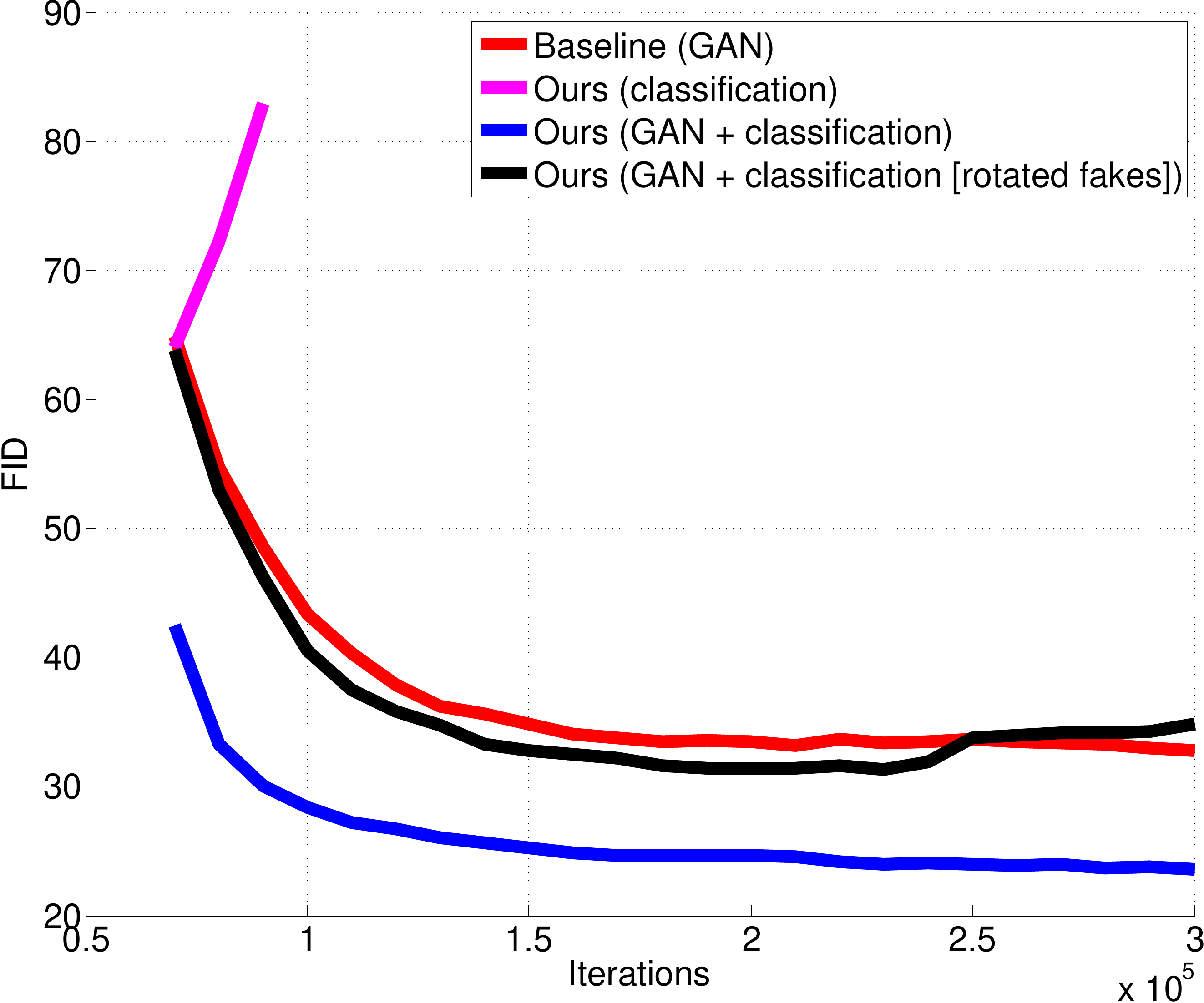}
\caption{The ablation study with DCGAN architecture on CIFAR-10 dataset. (a) Fine-tuning $\lambda_d$ of discriminator (without adversarial training). In this experiment, we set $\lambda_g = 0$. (b) Fine-tuning $\lambda_g$ of generator with $\lambda_d = 1.0$. We also train generator with a similar objective of SSGAN \cite{chen-arxiv-2018} with and without adversarial training for the discriminator. (c) Our experiment on the combination of GAN and classification tasks. When we remove the GAN task in our model and rotate the fake samples when training the discriminator, they get collapsed or decreased the quality of generated samples. We set $\lambda_d = 1.0$, $\lambda_g = 0.1$ for this experiment.}
\label{ss_finetuning_dcgan}
\end{figure*}


At first, we aim to seek good $\lambda_d$ and $\lambda_g$ of our proposed method. However, estimating both at the same time is expensive. Therefore, we propose to first seek the good $\lambda_d$ of the classification task for the discriminator (Eq. \ref{eq_D_obj}). We train the classification task of the discriminator with only the real transformed samples like \cite{chen-arxiv-2018}. We follow the geometric transformation of \cite{gidaris-iclr-2018}, which is simple but effective and achieved the best performance in self-supervised tasks, to argument images and their pseudo labels. In particular, we train discriminator to recognize the 2D rotations which are applied to the input image. We rotate the input image with $K = 4$ rotations ($0^{\circ}, 90^{\circ}, 180^{\circ}, 270^{\circ}$) and assign them the pseudo-labels from 1 to $K$. In this experiment, we set $\lambda_g = 0$. This ablation study is with DCGAN on CIFAR-10. Fig. \ref{ss_finetuning_dcgan}a shows that training discriminator with the self-supervised learning task stabilizes the baseline and makes the model converging faster. This technique helps to improve FID score from our baseline. This study suggests the good $\lambda_d = 1.0$ for DCGAN architecture.

The second study seeks a good $\lambda_g$ of the self-supervised task for the generator (Eq. \ref{eq:genobj_new}). Experimental conditions are exactly the same as the first one, excepts we fix the best $\lambda_d$ from the previous experiment. We consider the version with the best $\lambda_d$ from the previous study as the self-supervised baseline (SS). First, we investigate the influence of our self-supervised task for the generator on the overall performance. For that, we train the classification task of the discriminator without adversarial training (no fake class) as the previous experiment and train the generator for two cases: the similar objective of SSGAN \cite{chen-cvpr-2011} and our proposed objective in Eq. \ref{entropy_matching_loss}. We carry out the investigation with DCGAN architecture on CIFAR-10 dataset as shown in Fig. \ref{ss_finetuning_dcgan}b. The results show that training our generator with a similar objective of \cite{chen-cvpr-2011} causes the divergence issue and this generator objective does not help to improve the performance. However, when we use our proposed generator objective (Eq. \ref{eq:genobj_new}), the performance ($\lambda_g = 0.1$) is better than the self-supervised baseline. This confirms the usefulness of our proposed generator objective.

\begin{figure*}
\centering
\includegraphics[width=4cm,height=3.5cm]{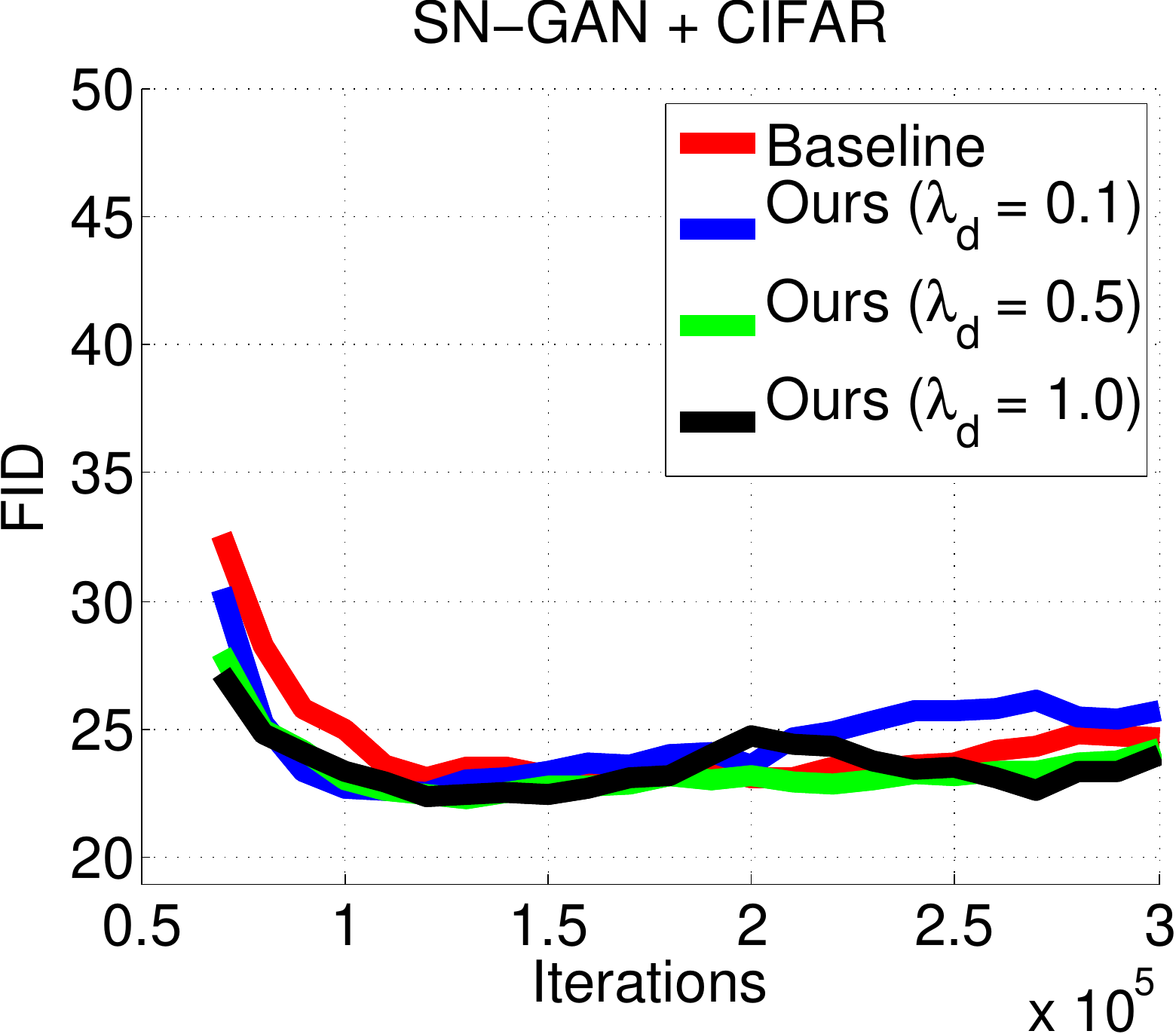}
\includegraphics[width=4cm,height=3.5cm]{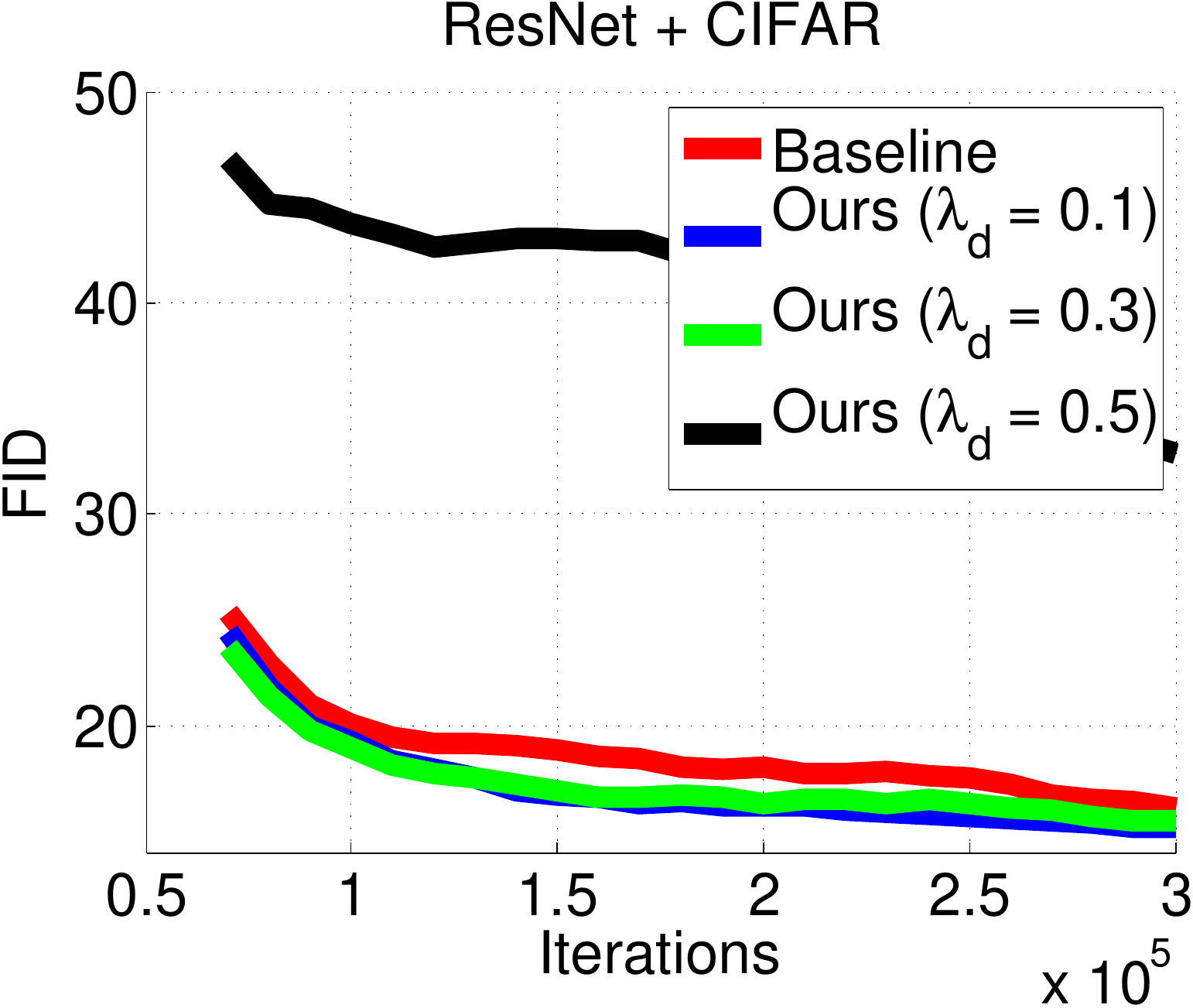}
\includegraphics[width=4cm,height=3.5cm]{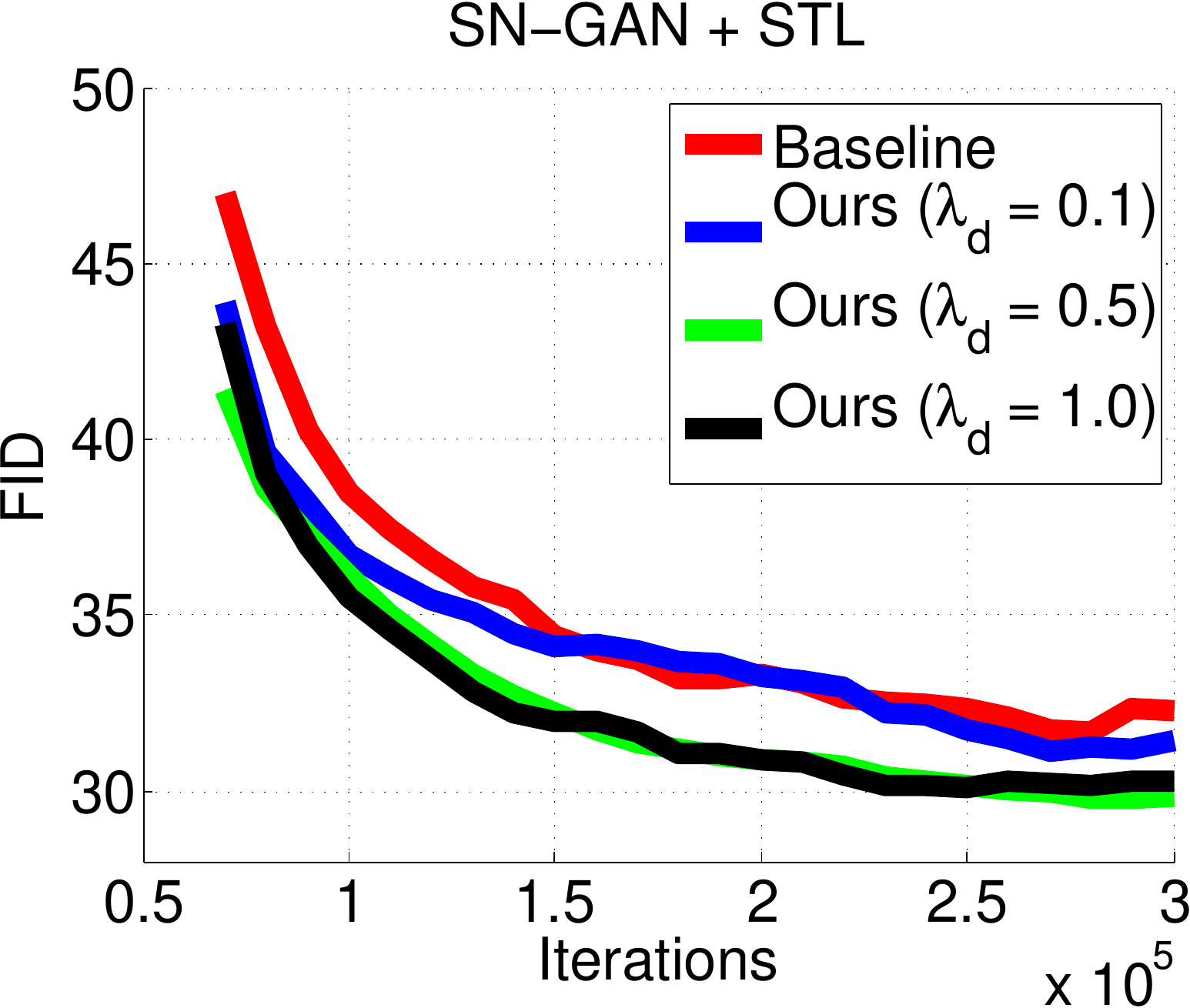}
\includegraphics[width=4cm,height=3.5cm]{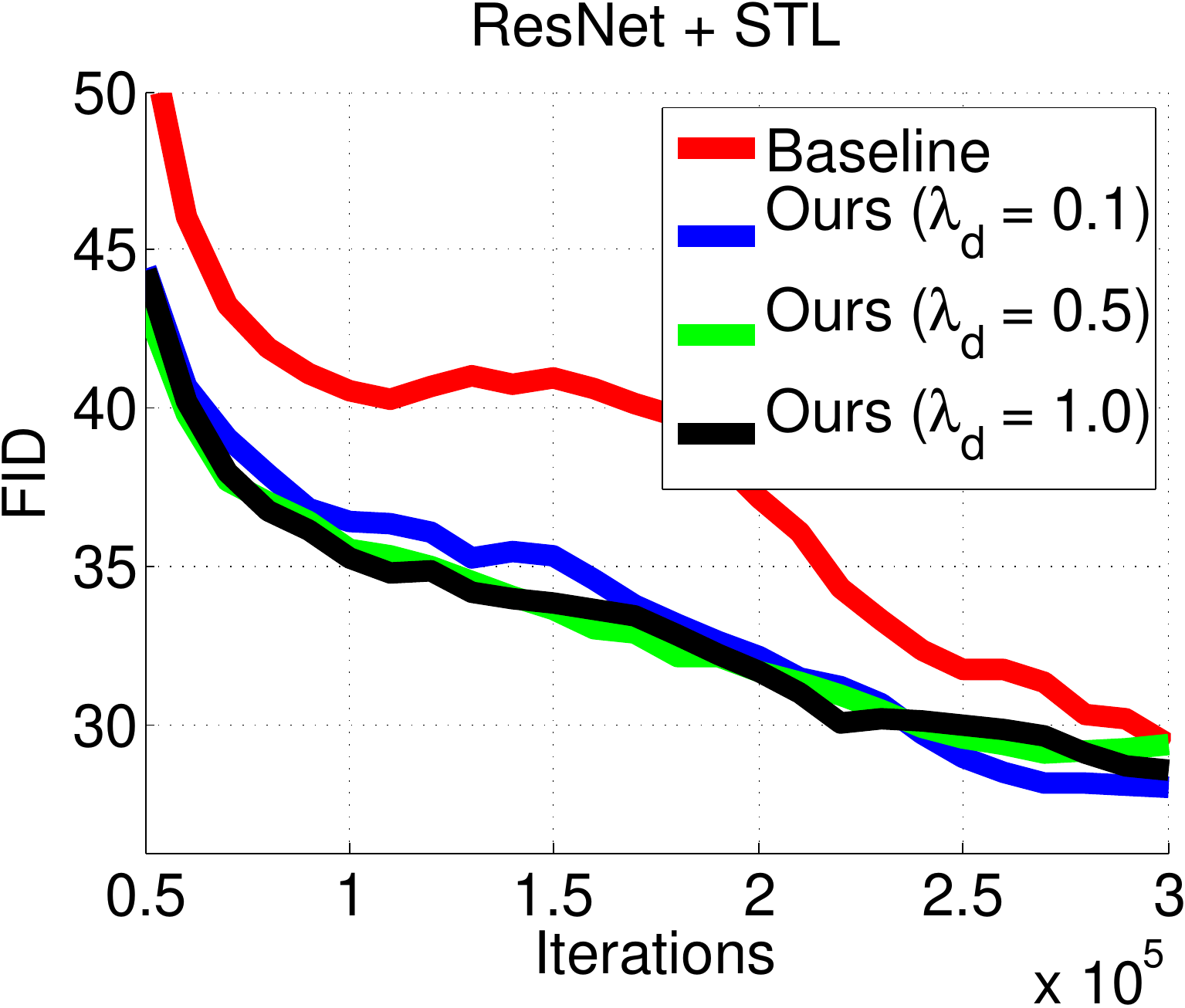}
\includegraphics[width=4cm,height=3.3cm]{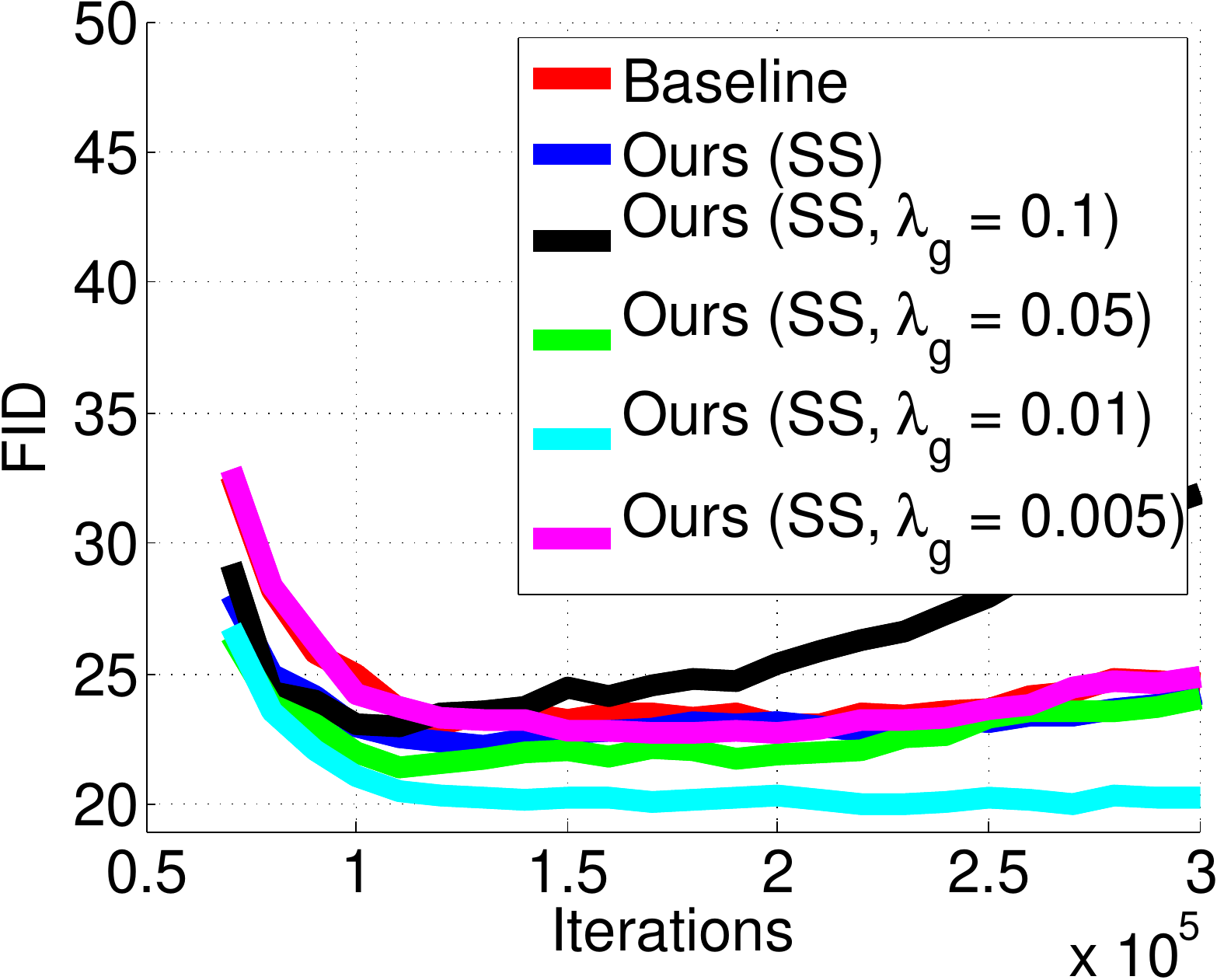}
\includegraphics[width=4cm,height=3.3cm]{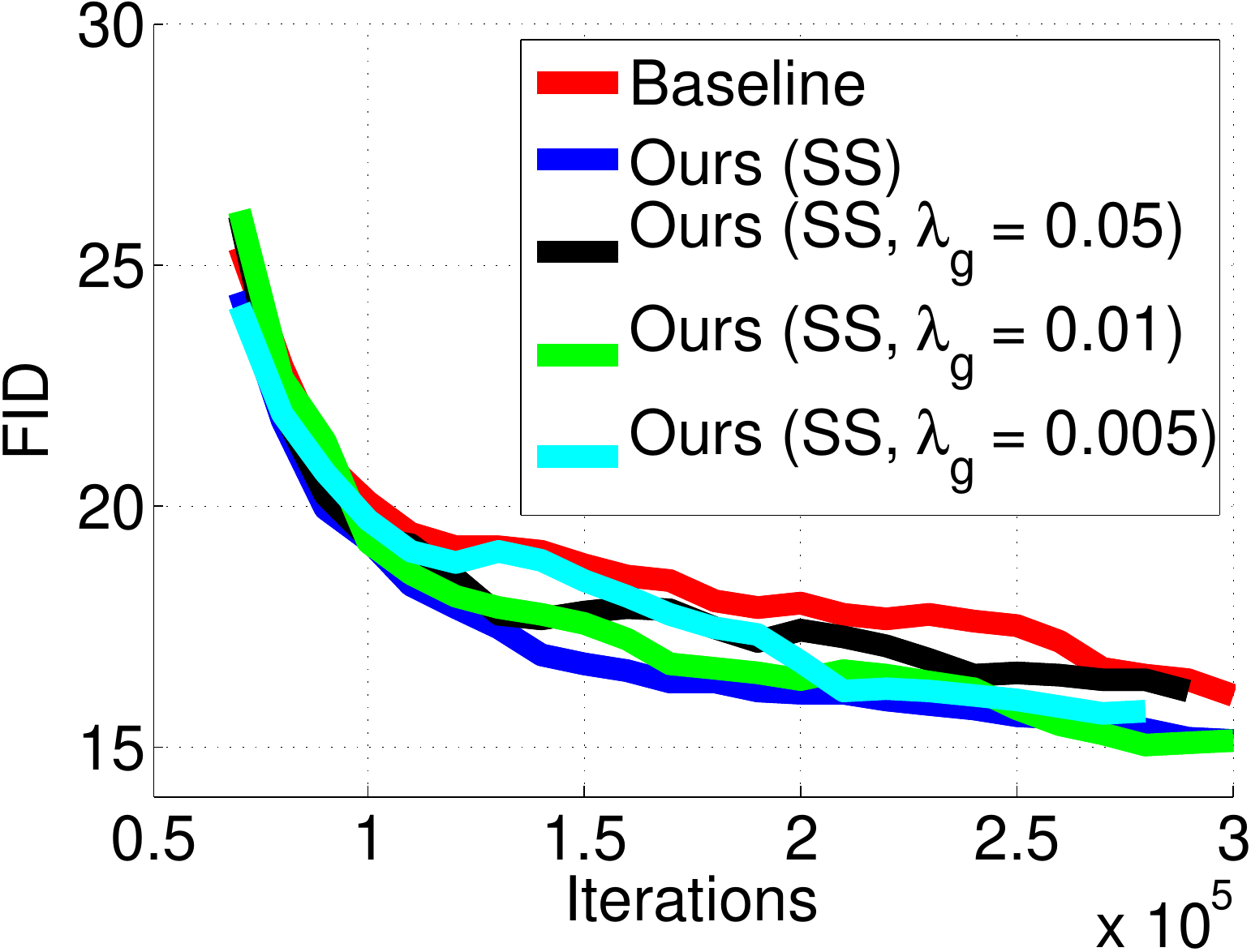}
\includegraphics[width=4cm,height=3.3cm]{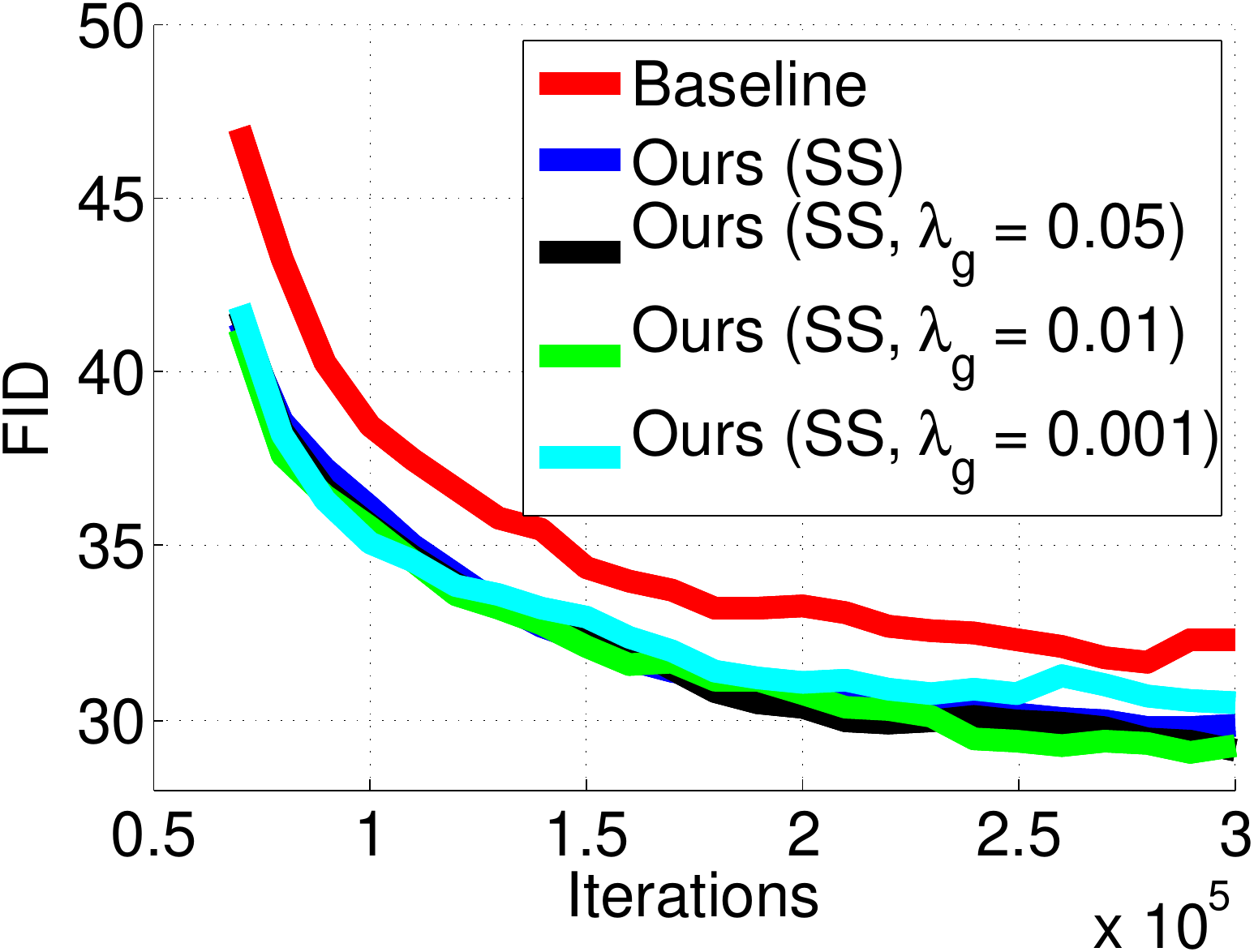}
\includegraphics[width=4cm,height=3.3cm]{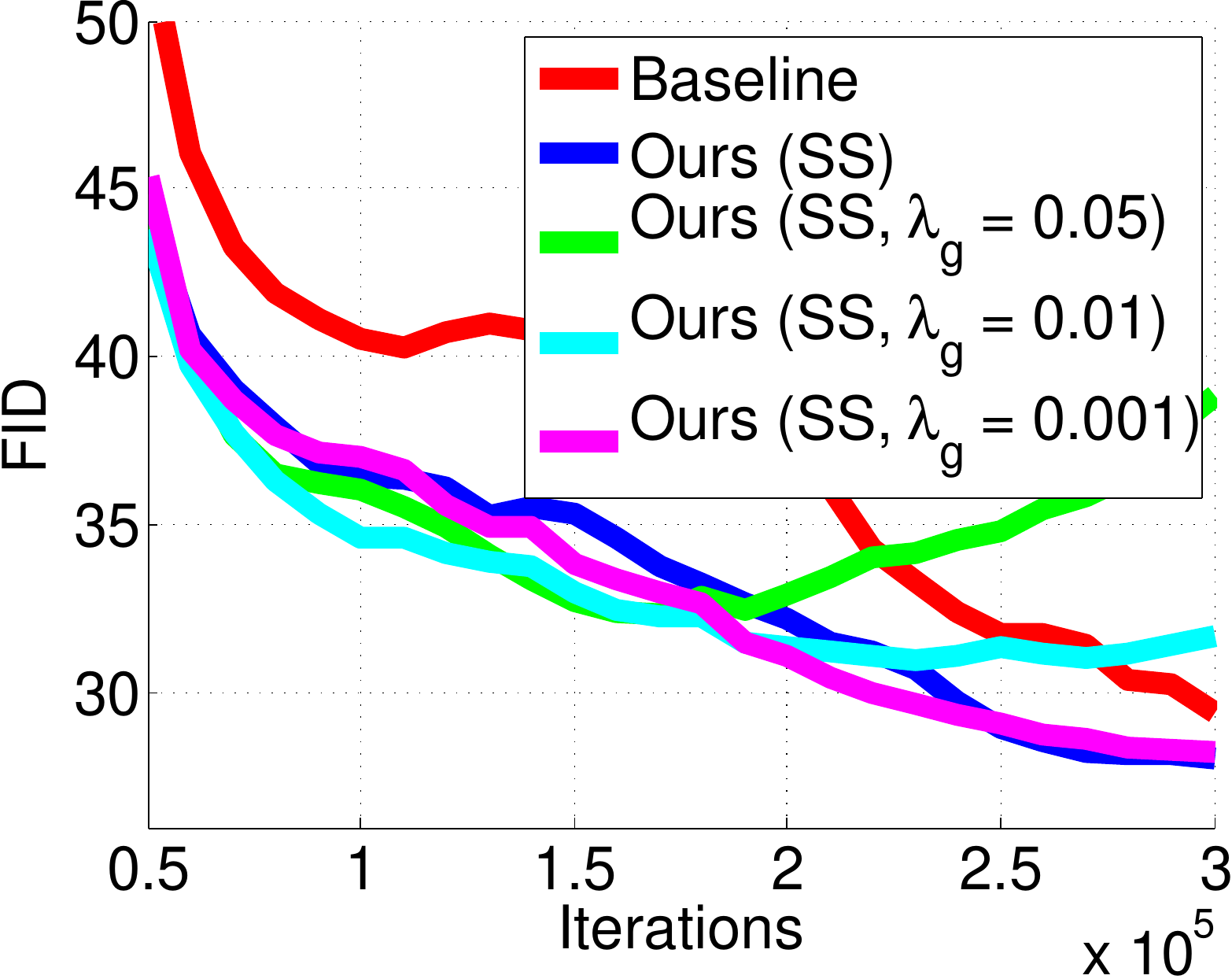}
\caption{The ablation study with SN-GAN and ResNet architectures on CIFAR-10 and STL-10 datasets. First row is fine-tuning $\lambda_d$ and second row is fine-tuning $\lambda_g$. From first to fourth columns: SN-GAN on CIFAR-10, ResNet on CIFAR-10, SN-GAN on STL-10 and ResNet on STL-10. The results suggest $\lambda_d = 0.5$ for SN-GAN and $\lambda_d = 0.1$ for ResNet. SS is the self-supervised baseline (with best $\lambda_d$). Adversarial training is applied for the discriminator when $\lambda_g > 0$.}
\label{ss_finetuning_others}
\end{figure*}

\begin{table*}
\centering
\caption{Comparing our best FID scores to the state of the art (Smaller is better). Methods with the SN-GAN \cite{miyato-iclr-2018} and ResNet (R) \cite{gulrajani-arxiv-2017,miyato-iclr-2018} architectures. FID scores of SN-GAN, Dist-GAN and our method reported with hinge loss. Performance of compared methods are from \cite{miyato-iclr-2018,tran-aaai-2018}. We also compare to SS-GAN \cite{chen-arxiv-2018} with the same 10K-10K FID scores for CIFAR-10 dataset with ResNet (R). (SS + adversarial, G) is with adversarial training and the new generator objective. 
}
\begin{tabular}{ c | c | c | c | c | c}
        \textbf{Method} & \textbf{CIFAR-10} & \textbf{STL-10} & \textbf{CIFAR-10 (R)} & \textbf{STL-10 (R)} & \textbf{CIFAR-10 (R) (10K-10K)}\\
\hline
\hline
GAN-GP \cite{miyato-iclr-2018}    			      & 37.7   & -      & - & - & - \\
WGAN-GP \cite{miyato-iclr-2018}    			      & 40.2   & 55.1   & - & - & - \\
SN-GAN \cite{miyato-iclr-2018}    			      & 25.5   & 43.2   & 21.70 $\pm$ .21 & 40.10 $\pm$ .50  & 19.73 \\
SS-GAN \cite{chen-arxiv-2018}                     & -      & -      & -               & -                & 15.65 \\
Dist-GAN \cite{tran-eccv-2018}  			      & 22.95  & 36.19  & 17.61 $\pm$ .30 & 28.50 $\pm$ .49  & 13.01 \\
GN-GAN \cite{tran-aaai-2018}                      & 21.70  & 30.80  & 16.47 $\pm$ .28 & -  & - \\
\hline
\textbf{Ours (SS)}                           & 21.40  & 29.79  & 14.97 $\pm$ .29 & \textbf{27.98 $\pm$ .38} & 12.37 \\   
\textbf{Ours (SS + adversarial, G)}                   & \textbf{19.05} & \textbf{28.70} & \textbf{14.75 $\pm$ .28} & 28.24 $\pm$ .23 & \textbf{12.15}  \\
\end{tabular}
\label{fid_score}
\end{table*}

Third, we want to understand the influence of adversarial training (with a fake class) for the classification task given best $\lambda_d$, $\lambda_g$ from previous studies. This experiment is also with DCGAN architecture on CIFAR-10 dataset (Fig. \ref{ss_finetuning_dcgan}b). We now train the discriminator with adversarial learning (simultaneously distinguishing fake class in the classification task). Note in our experiments, when we mention about adversarial training, we mean using it for the classification task. We first carry out experiments with our proposed generator objective. When considering the adversarial training (with the fake class) for the discriminator, our method improves FID significantly as comparing the non-adversarial version. We also try with the generator objective of \cite{chen-cvpr-2011}. Although FID is slightly improved from the self-supervised baseline, using this generator objective still gets collapsed. In contrast, our proposed objective is stable and achieve the best FID than other versions (Fig. \ref{ss_finetuning_dcgan}b). This confirms the importance of the combination of our adversarial self-supervised learning and our proposed generator objective. The reason of training generator objective like \cite{chen-cvpr-2011} leads to the corruption is perhaps because of maximizing it violates the GAN task of our generator, which does not support the match of D(x) and D(G(z)) of the first term. This violation is similar to the gradient penalty \cite{gulrajani-arxiv-2017} although it may be useful at the beginning but diverge at the end. Intuitively, our new objective (Eq. \ref{entropy_matching_loss}) does not violate because when data and generator distributions are matched, their classification should be similar either. This study again verifies the hypothesis of our proposed techniques.

Fourth, in previous experiments, we figured out how the classification task helps to improve the GAN task. Although seemly there is overlapping between GAN task and classification task as they both classify the same fake sample, having both tasks in the model is important. For instance, if removing the GAN task in our model (for both discriminator and generator), the model gets immediately collapsed at first iterations as shown in Fig. \ref{ss_finetuning_dcgan}c. It means that the GAN task still plays an important role in our GAN model. We also consider the adversarial training of discriminator objective like (Eq. \ref{obj_classification}) but we now rotate the fake samples and consider these rotated samples belonging to the fake class. The result in Fig. \ref{ss_finetuning_dcgan}c does not suggest to rotate the fake samples when training the discriminator, because it likely creates noise and degrades its learning. We conduct this study with DCGAN on CIFAR-10 in the similar experimental setup previous studies, and $\lambda_d = 1.0$ and $\lambda_g = 0.1$ if the classification task is used.

Fifth, we also investigate proposed techniques for other network architectures CIFAR-10 and STL-10 datasets. At first, we repeat the first experiment for SN-GAN and ResNet architectures to select their best $\lambda_d$ as shown in the first row of Fig. \ref{ss_finetuning_others}. The results suggest $\lambda_d = 0.5$ for SN-GAN and $\lambda_d = 0.1$ for Resnet. We realize that when the network is powerful (eg. ResNet), the best $\lambda_d$ gets smaller. Perhaps, the more powerful network has better capability to learn good feature representation via the GAN task. In contrast, the smaller networks (DCGAN, SN-GAN) are harder to train, therefore needs more contribution from the classification task. Then, we study the good $\lambda_g$ for these architectures as shown in the second row of Fig. \ref{ss_finetuning_others}. Here, we seek $\lambda_g$ for our generator objective in the case of that the classifier is trained with the fake class (adversarial training) similar to our third study with DCGAN architecture. The generator objective helps to boost significantly the performance (if the choice of $\lambda_g$ is good), especially for SN-GAN on CIFAR-10. Our proposed techniques also reduce the divergence issue as shown in the first column of Fig. \ref{ss_finetuning_others}. Although the baseline with ResNet achieves almost saturated performance, our techniques are still able to improve this model further. It's worth noting that the FID of our self-supervised baseline (SS) already reaches the similar performance of SAGAN \cite{zhang-arxiv-2018} - the state-of-the-art conditional GAN (see the discussion in the supplementary material) - it's hard to make the improvement higher even though being combined with the adversarial training and our proposed generator objective. This study again confirms the effectiveness and robustness of our proposed techniques on various architecture. We observe that $\lambda_g \sim \lambda_d/10$ are good choices for DCGAN and ResNet on CIFAR-10, and $\lambda_g \sim \lambda_d/50$ is good for SN-GAN on CIFAR-10 and STL-10. With the combination of adversarial self-supervised learning for discriminator and our proposed generator objective, our best versions significantly outperform the baseline for various network architectures and datasets.

\subsection{Compared to state-of-the-art methods}

\begin{figure}
\centering
\includegraphics[scale=0.45]{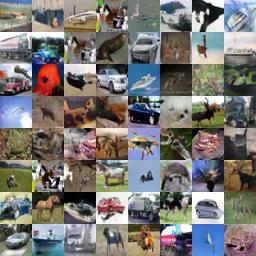}
\includegraphics[scale=0.30]{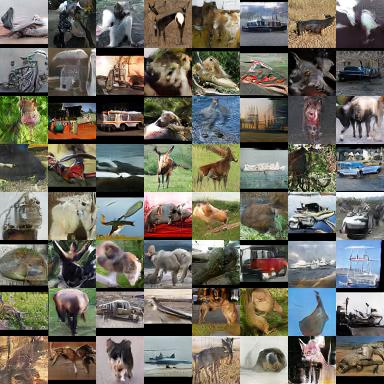}
\caption{Samples are generated by our method on CIFAR-10 and STL-10 datasets with ResNet architectures.}
\label{fig:examples}
\end{figure}

In this section, we compare the best settings of our proposed method to the state of the art on benchmark datasets: CIFAR-10 and STL-10 as shown in Table \ref{fid_score}. We compare results obtained with SN-GAN \cite{miyato-iclr-2018} and ResNet \cite{gulrajani-arxiv-2017,miyato-iclr-2018} architectures. As shown in Table \ref{fid_score}, our method significantly outperforms the baseline Dist-GAN and other GAN methods, especially on the STL-10 dataset. This confirms the effectiveness of the combination GAN task and classification task into a unique model. It's worth-noting that SN-GAN attains best results at about 100K iterations, yet this model diverges if continue the training. The similar observation is also discussed in \cite{chen-arxiv-2018}. We also compare to the recent work, SSGAN \cite{chen-arxiv-2018}, which also integrates the self-supervised technique to improve GAN model. For this case, to be a fair comparison, we compute the similar FID with 10K real samples and 10K fake samples like this work. Our model achieves much better FID score than SSGAN with same ResNet architecture on CIFAR-10 dataset. Fig. \ref{fig:examples} show some generated examples of our model of ResNet architectures on CIFAR-10 and STL-10 datasets.

\section{Conclusion}

We propose to train the model with adversarial self-supervised learning. First, we show that training self-supervised learning helps to improve the discriminator (self-supervised baseline) and hence enhance the quality of generated images. Then we propose to train the discriminator with adversarial and a new generator objective via matching the cross-entropy loss between real and fake samples. The combination of adversarial training (discriminator) and cross-entropy matching (for generator) further boosts the performance of self-supervised baseline over with various network architectures on CIFAR-10 and STL-10 datasets. The best version of our proposed method significantly outperformed the baseline and established the new state-of-the-art FID scores over these benchmark datasets. Although investigating our proposed techniques mainly within an auto-encoder GAN model, we believe that our proposed techniques are orthogonal and potential to be used to improve other GAN methods.

{\small
\bibliographystyle{ieee}
\bibliography{../../../biblio/biblio}
}

\end{document}